\documentclass[conference]{IEEEtran}

\usepackage[utf8]{inputenc}
\usepackage[T1]{fontenc}
\usepackage{cite}
\usepackage{amsmath,amssymb,amsfonts}
\usepackage{algorithmic}
\usepackage{algorithm}
\usepackage{graphicx}
\usepackage{textcomp}
\usepackage{xcolor}
\usepackage{booktabs}
\usepackage{multirow}
\usepackage{hyperref}
\usepackage{url}
\usepackage{listings}
\usepackage{subcaption}
\usepackage{bm}

\newcommand{\E}{\mathbb{E}}

\newcommand{\N}{\mathcal{N}}
\newcommand{\Loss}{\mathcal{L}}

\lstdefinestyle{python}{
    language=Python,
    basicstyle=\ttfamily\scriptsize,
    keywordstyle=\color{blue},
    stringstyle=\color{red},
    commentstyle=\color{green!60!black},
    numbers=left,
    numberstyle=\tiny\color{gray},
    stepnumber=1,
    numbersep=5pt,
    backgroundcolor=\color{gray!10},
    showspaces=false,
    showstringspaces=false,
    frame=single,
    tabsize=2,
    breaklines=true,
}

\def\BibTeX{{\rm B\kern-.05em{\sc i\kern-.025em b}\kern-.08em
    T\kern-.1667em\lower.7ex\hbox{E}\kern-.125emX}}

\begin{document}

\title{From Diffusion to One-Step Generation: A Comparative Study of Flow-Based Models with Application to Image Inpainting}

\author{
\IEEEauthorblockN{Umang Agarwal}
\IEEEauthorblockA{\textit{Department of Electrical Engineering} \\
\textit{IIT Bombay}\\
21D070083 \\
umanga@iitb.ac.in}
\and
\IEEEauthorblockN{Rudraksh Sangore}
\IEEEauthorblockA{\textit{Department of Electrical Engineering} \\
\textit{IIT Bombay}\\
21D070060 \\
rudrakshsangore@iitb.ac.in}
\and
\IEEEauthorblockN{Sumit Laddha}
\IEEEauthorblockA{\textit{Department of Electrical Engineering} \\
\textit{IIT Bombay}\\
21D070077 \\
21d070077@iitb.ac.in}
}

\maketitle

\begin{abstract}
We present a comprehensive comparative study of three generative modeling paradigms: Denoising Diffusion Probabilistic Models (DDPM), Conditional Flow Matching (CFM), and MeanFlow. While DDPM and CFM require iterative sampling, MeanFlow enables direct one-step generation by modeling the average velocity over time intervals. We implement all three methods using a unified TinyUNet architecture ($<$1.5M parameters) on CIFAR-10, demonstrating that CFM achieves an FID of 24.15 with 50 steps, significantly outperforming DDPM (FID 402.98). MeanFlow achieves FID 29.15 with single-step sampling---a 50$\times$ reduction in inference time. We further extend CFM to image inpainting, implementing mask-guided sampling with four mask types (center, random bbox, irregular, half). Our fine-tuned inpainting model achieves substantial improvements: PSNR increases from 4.95 to 8.57 dB on center masks (+73\%), and SSIM improves from 0.289 to 0.418 (+45\%), demonstrating the effectiveness of inpainting-aware training. Code is available at \url{https://github.com/AgarwalUmang/aml_flow_matching_project}.
\end{abstract}


\section{Introduction}
\label{sec:introduction}

Generative modeling has witnessed remarkable progress with diffusion-based methods \cite{ho2020denoising, song2020score} and flow matching techniques \cite{lipman2023flow, liu2023flow}. These approaches have achieved state-of-the-art results in image synthesis, but typically require many sampling steps during inference.

This work presents a unified study of three generative paradigms with progressively improving sampling efficiency, plus an extension to image inpainting:

\textbf{DDPM} \cite{ho2020denoising} learns to reverse a gradual noising process through iterative denoising. Our experiments show that with limited training, DDPM achieves FID of 402.98.

\textbf{Conditional Flow Matching (CFM)} \cite{lipman2023flow} learns a velocity field via ODE integration, achieving FID of 24.15---a 16.7$\times$ improvement over DDPM.

\textbf{MeanFlow} \cite{meanflow2025} models the average velocity, enabling \textbf{one-step sampling} with FID of 29.15.

\textbf{CFM Inpainting:} We extend CFM to mask-guided image inpainting, demonstrating that fine-tuning improves PSNR by 73\% (4.95 $\to$ 8.57 dB).

\subsection{Contributions}
\begin{itemize}
    \item Unified implementations of DDPM, CFM, and MeanFlow using consistent TinyUNet architectures
    \item Comprehensive evaluation with FID, KID, and per-class metrics
    \item CFM-based inpainting with four mask types and detailed per-class analysis
    \item Fine-tuning strategy that improves inpainting PSNR by 73\% on average
    \item Open-source code with training and evaluation scripts
\end{itemize}

\section{Denoising Diffusion Probabilistic Models}
\label{sec:ddpm}

\subsection{Forward Process with Cosine Schedule}
DDPM defines a forward noising process with cosine schedule \cite{nichol2021improved}:
\begin{equation}
    q(x_t | x_0) = \N(x_t; \sqrt{\bar{\alpha}_t} x_0, (1 - \bar{\alpha}_t) \mathbf{I})
\end{equation}
where $\bar{\alpha}_t = f(t)/f(0)$ with $f(t) = \cos^2((t/T + s)/(1 + s) \cdot \pi/2)$, $s = 0.008$, and $T = 200$.

\subsection{Training Objective}
\begin{equation}
    \Loss_{\text{DDPM}} = \E_{t, x_0, \epsilon}\left[\|\epsilon - \epsilon_\theta(x_t, t/T, y)\|^2\right]
\end{equation}

\subsection{DDIM Sampling}
We use DDIM \cite{song2020denoising} for deterministic sampling:
\begin{align}
    \hat{x}_0 &= \frac{x_t - \sqrt{1-\bar{\alpha}_t} \epsilon_\theta}{\sqrt{\bar{\alpha}_t}} \\
    x_{t-1} &= \sqrt{\bar{\alpha}_{t-1}} \hat{x}_0 + \sqrt{1-\bar{\alpha}_{t-1}} \epsilon_\theta
\end{align}

\section{Conditional Flow Matching}
\label{sec:cfm}

\subsection{Rectified Flow Formulation}
CFM uses linear optimal transport interpolation:
\begin{equation}
    x_t = (1-t) x_0 + t x_1, \quad t \in [0, 1]
\end{equation}
where $x_0 \sim \N(0, \mathbf{I})$ is noise and $x_1 \sim q(x_1)$ is data.

\subsection{Training Objective}
The target velocity is constant along the path: $v_{\text{target}} = x_1 - x_0$:
\begin{equation}
    \Loss_{\text{CFM}} = \E_{t, x_0, x_1}\left[\|v_\theta(x_t, t, y) - (x_1 - x_0)\|^2\right]
\end{equation}

\subsection{Sampling with Classifier-Free Guidance}
CFM samples via Euler integration with classifier-free guidance:
\begin{equation}
    \tilde{v} = v_\theta(x_t, t, \varnothing) + w \cdot (v_\theta(x_t, t, y) - v_\theta(x_t, t, \varnothing))
\end{equation}

\section{MeanFlow: One-Step Generation}
\label{sec:meanflow}

\subsection{Average Velocity Formulation}
MeanFlow models the average velocity over interval $[r, t]$:
\begin{equation}
    u_\theta(z, r, t) \approx \frac{1}{t-r} \int_r^t v(z_s, s) ds
\end{equation}

\subsection{MeanFlow Identity}
The training target uses JVP (Jacobian-Vector Product) computation:
\begin{equation}
    u_{\text{target}} = v_t - (t-r) \cdot \left(v_t \cdot \nabla_z u_\theta + \frac{\partial u_\theta}{\partial t}\right)
\end{equation}

\subsection{One-Step Sampling}
The key advantage is direct generation without ODE integration:
\begin{equation}
    x = \varepsilon - u_\theta(\varepsilon, 0, 1, y)
\end{equation}

\section{CFM-Based Image Inpainting}
\label{sec:inpainting}

We extend Conditional Flow Matching to image inpainting through mask-guided sampling. Given an image $x$ and binary mask $m$ (where $m=1$ indicates known regions and $m=0$ indicates regions to inpaint), we generate realistic completions while preserving the known content.

\subsection{Mask Types}
We implement four mask generation strategies (Fig.~\ref{fig:mask_radar}):

\textbf{Center Mask:} A 16$\times$16 square centered in the image, covering 25\% of pixels.

\textbf{Random Bounding Box:} Random rectangular regions with dimensions 8--20 pixels.

\textbf{Irregular Brush Strokes:} 3--8 random brush strokes with varying angles and widths.

\textbf{Half Image:} Removes entire left/right/top/bottom half---the most challenging case.

\subsection{Mask-Guided Sampling Algorithm}
The inpainting algorithm modifies CFM sampling to constrain the known regions:

\begin{algorithm}[t]
\caption{CFM Inpainting (Replace Strategy)}
\label{alg:inpainting}
\begin{algorithmic}[1]
\REQUIRE Image $x$, mask $m$, model $v_\theta$, steps $N$
\STATE $z \leftarrow m \cdot x + (1-m) \cdot \epsilon$, $\epsilon \sim \N(0, I)$
\FOR{$i = 0$ to $N-1$}
    \STATE $t \leftarrow i/N$
    \STATE $v \leftarrow v_\theta(z, t, y)$ with CFG
    \STATE $z \leftarrow z + (1/N) \cdot v$
    \STATE $z \leftarrow m \cdot x + (1-m) \cdot z$ \COMMENT{Replace known regions}
\ENDFOR
\RETURN $z$
\end{algorithmic}
\end{algorithm}

\subsection{Inpainting-Aware Fine-Tuning}
We introduce a fine-tuning strategy that trains the model with masked inputs:
\begin{equation}
    \Loss_{\text{inpaint}} = 0.7 \cdot \Loss_{\text{full}} + 0.3 \cdot \Loss_{\text{masked}}
\end{equation}
where $\Loss_{\text{full}} = \|v_\theta(\tilde{x}_t, t, y) - v_{\text{target}}\|^2$ and $\Loss_{\text{masked}} = \|(v_\theta - v_{\text{target}}) \cdot (1-m)\|^2$.

During training, we modify the interpolated sample: $\tilde{x}_t = m \cdot x_1 + (1-m) \cdot x_t$, simulating the inference-time conditioning.

\subsection{Evaluation Metrics}
We evaluate inpainting quality on the masked regions using:

\textbf{NMSE} (Normalized Mean Squared Error): $\text{NMSE} = \frac{\sum_{i \in M} (x_i - \hat{x}_i)^2}{\sum_{i \in M} x_i^2}$

\textbf{PSNR} (Peak Signal-to-Noise Ratio): $\text{PSNR} = 10 \log_{10}(\text{MAX}^2/\text{MSE})$

\textbf{SSIM} (Structural Similarity Index) \cite{wang2004image}

\section{Experiments}
\label{sec:experiments}

\subsection{Experimental Setup}

\textbf{Dataset:} CIFAR-10 (50K train, 10K test, 32$\times$32, 10 classes).

\textbf{Training Configuration:}
\begin{itemize}
    \item Optimizer: AdamW, lr=$3\times10^{-4}$, weight decay 0.01
    \item Scheduler: Cosine annealing to $10^{-4}$
    \item Epochs: 200 (CFM/MeanFlow), 400 (DDPM)
    \item Batch size: 128
    \item CFG dropout: 10\%, CFG scale: 2.0--3.0
\end{itemize}

\textbf{Inpainting Fine-tuning:}
\begin{itemize}
    \item Pre-trained: Best CFM checkpoint
    \item Fine-tune epochs: 20, Learning rate: $5\times10^{-5}$
    \item Mask probability: 50\% of batches
    \item CFG dropout: 15\%
\end{itemize}

\textbf{Evaluation:} 5,000 samples for generation (500/class), 5,000 for inpainting.

\subsection{Generation Results}

\begin{table}[t]
\centering
\caption{Overall Generation Performance (50 sampling steps)}
\label{tab:main_results}
\begin{tabular}{lcccc}
\toprule
\textbf{Method} & \textbf{FID}$\downarrow$ & \textbf{KID$\times$1000}$\downarrow$ & \textbf{NFE} & \textbf{Speed} \\
\midrule
CFM & \textbf{24.15} & \textbf{12.39} & 50 & 28.01 img/s \\
MeanFlow & 29.15 & 13.52 & \textbf{1} & \textbf{25.69 img/s} \\
DDPM & 402.98 & 463.21 & 50 & 27.39 img/s \\
\bottomrule
\end{tabular}
\end{table}

Table~\ref{tab:main_results} shows the overall generation metrics. Key findings:
\begin{itemize}
    \item CFM achieves 16.7$\times$ lower FID than DDPM (24.15 vs 402.98)
    \item MeanFlow achieves comparable FID with 50$\times$ fewer steps
    \item KID scores confirm the same ranking: CFM (12.39) $<$ MeanFlow (13.52) $\ll$ DDPM (463.21)
\end{itemize}

\begin{figure}[t]
\centering
\includegraphics[width=\columnwidth]{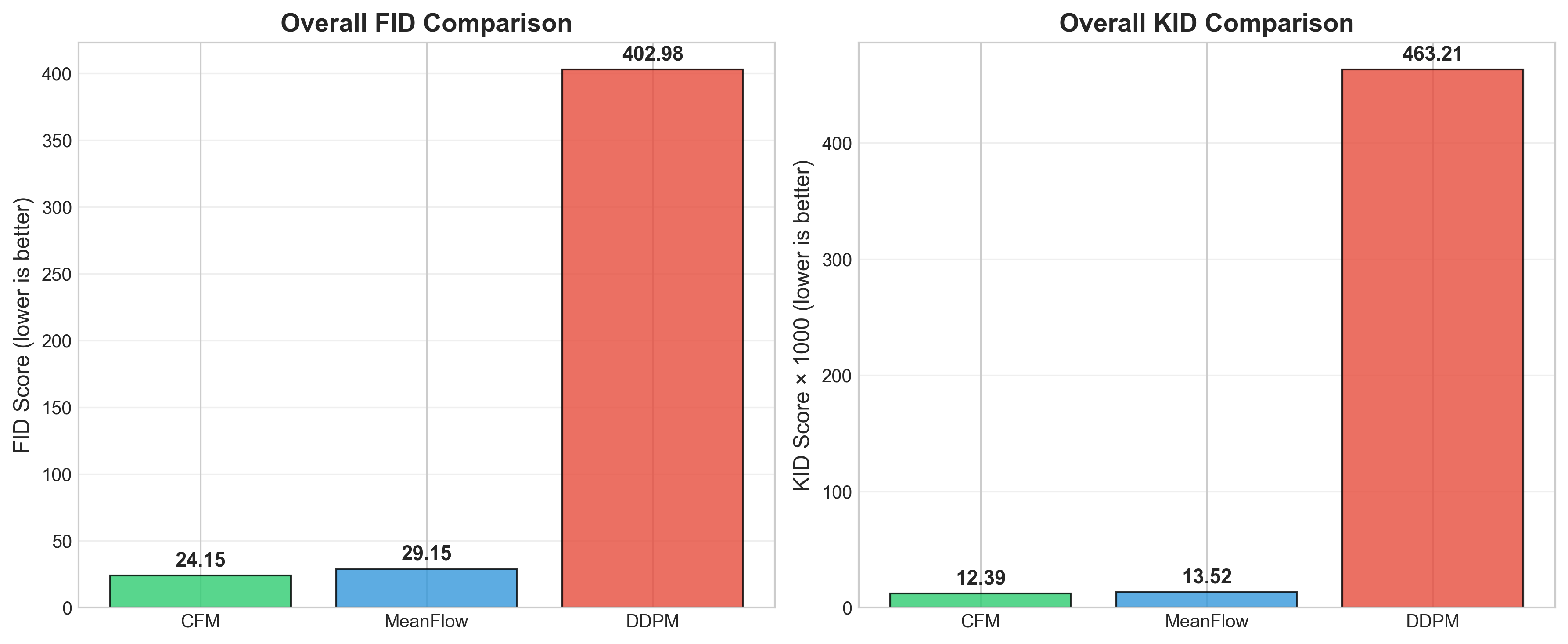}
\caption{Overall FID and KID comparison across three methods. CFM and MeanFlow significantly outperform DDPM, with CFM achieving the best scores.}
\label{fig:overall_metrics}
\end{figure}

\subsection{Qualitative Generation Results}

\begin{figure*}[t]
\centering
\begin{subfigure}[b]{0.24\textwidth}
    \includegraphics[width=\textwidth]{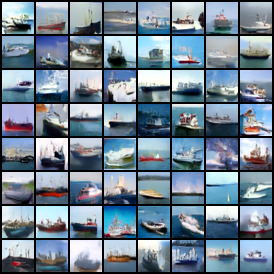}
    \caption{CFM: Ship (FID 50.38)}
\end{subfigure}
\hfill
\begin{subfigure}[b]{0.24\textwidth}
    \includegraphics[width=\textwidth]{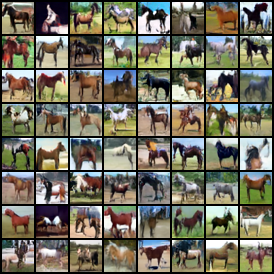}
    \caption{CFM: Horse (FID 57.58)}
\end{subfigure}
\hfill
\begin{subfigure}[b]{0.24\textwidth}
    \includegraphics[width=\textwidth]{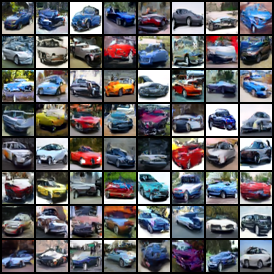}
    \caption{CFM: Automobile (FID 54.37)}
\end{subfigure}
\hfill
\begin{subfigure}[b]{0.24\textwidth}
    \includegraphics[width=\textwidth]{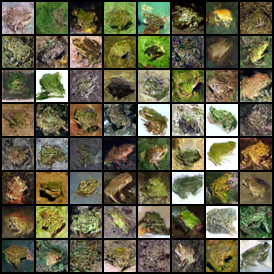}
    \caption{CFM: Frog (FID 71.41)}
\end{subfigure}
\caption{CFM generated samples (50 steps, CFG=3.0) for selected classes. Ship and automobile show clearest structure; frog demonstrates good texture capture.}
\label{fig:cfm_samples}
\end{figure*}

\begin{figure*}[t]
\centering
\begin{subfigure}[b]{0.24\textwidth}
    \includegraphics[width=\textwidth]{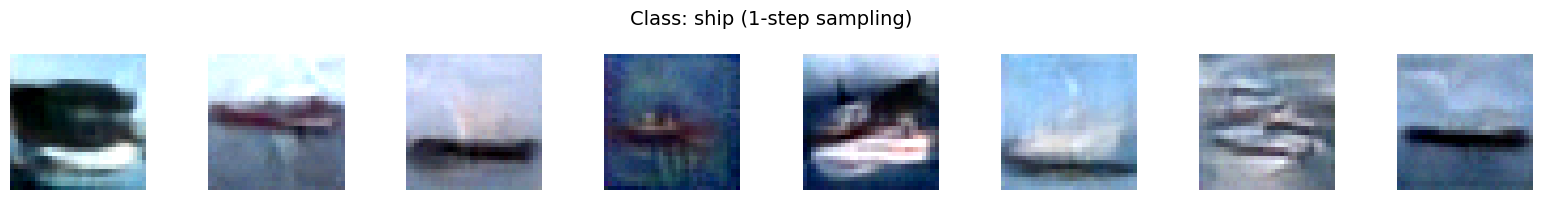}
    \caption{MeanFlow: Ship}
\end{subfigure}
\hfill
\begin{subfigure}[b]{0.24\textwidth}
    \includegraphics[width=\textwidth]{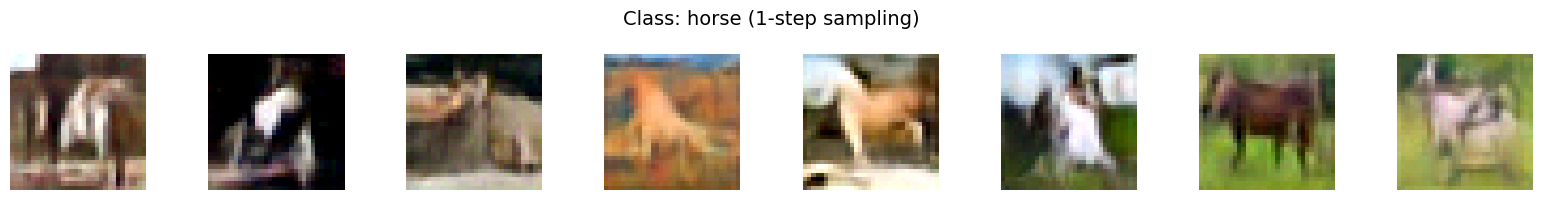}
    \caption{MeanFlow: Horse}
\end{subfigure}
\hfill
\begin{subfigure}[b]{0.24\textwidth}
    \includegraphics[width=\textwidth]{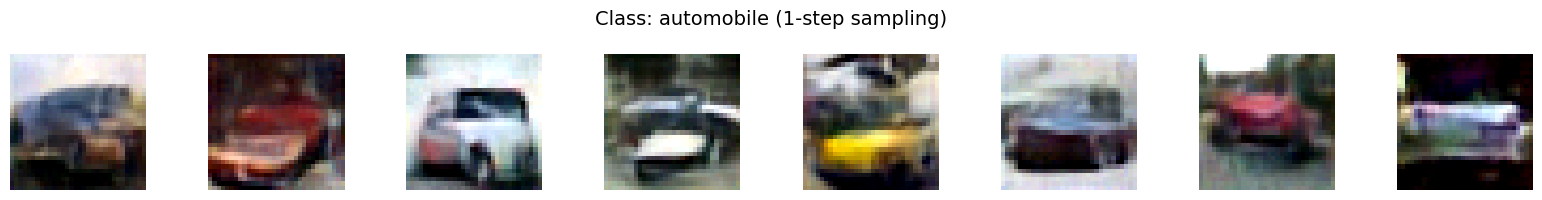}
    \caption{MeanFlow: Automobile}
\end{subfigure}
\hfill
\begin{subfigure}[b]{0.24\textwidth}
    \includegraphics[width=\textwidth]{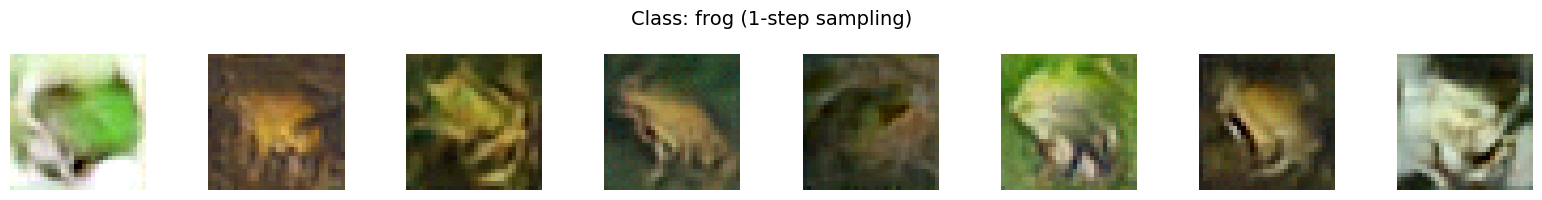}
    \caption{MeanFlow: Frog}
\end{subfigure}
\caption{MeanFlow generated samples with \textbf{single-step} sampling. Despite using only 1 NFE (vs 50 for CFM), samples maintain reasonable quality with recognizable objects and appropriate colors.}
\label{fig:meanflow_samples}
\end{figure*}

\begin{figure}[t]
\centering
\includegraphics[width=0.8\columnwidth]{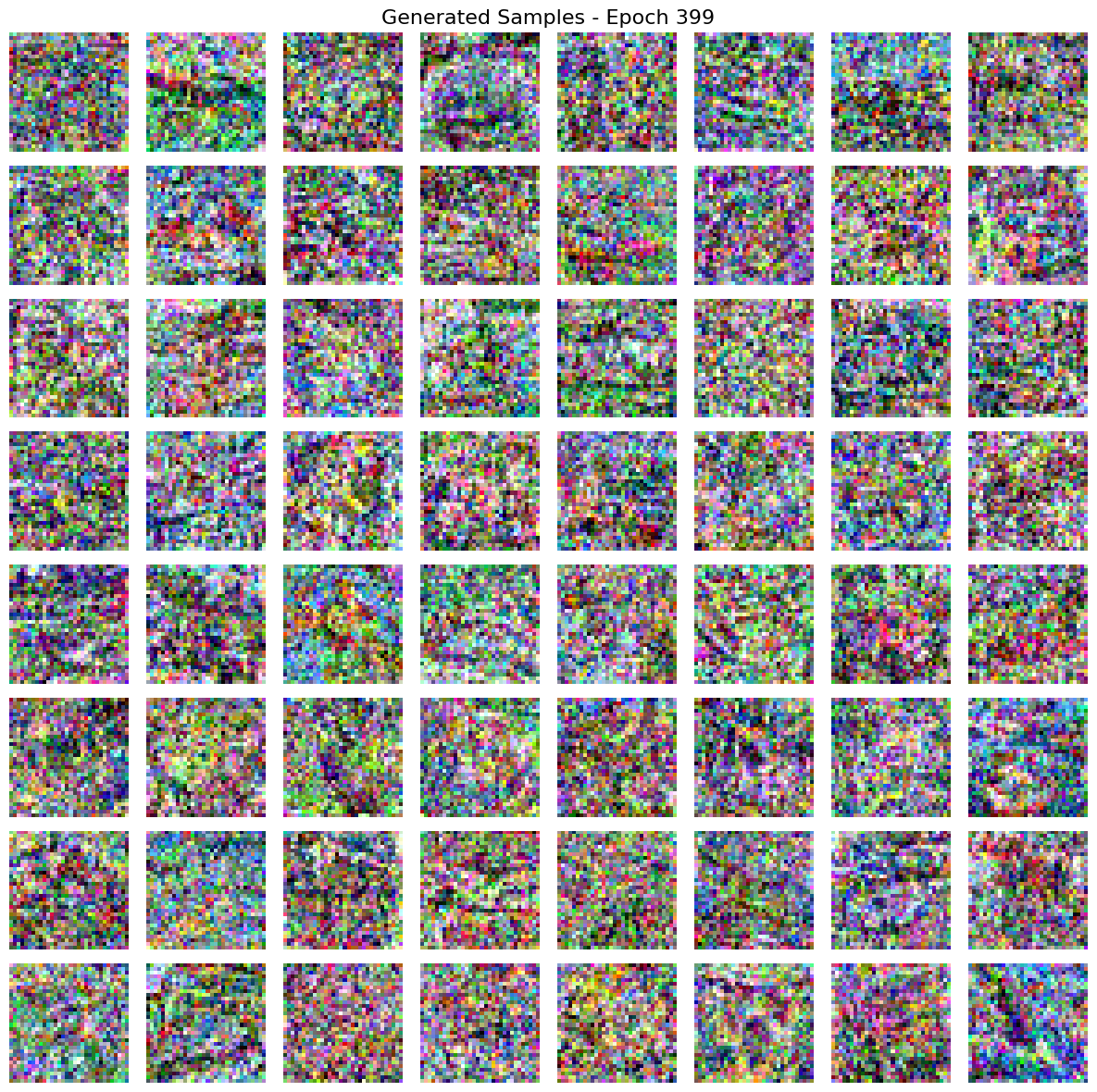}
\caption{DDPM samples at epoch 399. Despite extended training (400 epochs), the model fails to generate coherent images, producing only noise-like patterns. This explains the poor FID of 402.98.}
\label{fig:ddpm_failure}
\end{figure}

Fig.~\ref{fig:cfm_samples} shows CFM samples across different classes. The model captures class-specific features well, with ships showing clear blue backgrounds and vessels, horses displaying proper body structure, and automobiles maintaining recognizable shapes.

Fig.~\ref{fig:meanflow_samples} demonstrates MeanFlow's remarkable one-step generation capability. Despite using only a single forward pass, the samples show coherent objects with appropriate colors and textures, though with slightly less detail than CFM's 50-step samples.

Fig.~\ref{fig:ddpm_failure} illustrates DDPM's training failure. Even at epoch 399, samples remain noise-like, explaining the FID of 402.98. This failure is likely due to: (1) insufficient timesteps ($T=200$ vs typical $T=1000$), (2) suboptimal noise schedule for small networks, and (3) the inherent difficulty of noise prediction vs velocity matching.

\subsection{Per-Class Generation Analysis}

\begin{table}[t]
\centering
\caption{Per-Class FID Scores (50 steps)}
\label{tab:per_class_fid}
\begin{tabular}{lcc}
\toprule
\textbf{Class} & \textbf{CFM} & \textbf{MeanFlow} \\
\midrule
airplane & 65.13 & 74.85 \\
automobile & 54.37 & 63.94 \\
bird & 79.66 & 90.23 \\
cat & 75.63 & 86.18 \\
deer & 62.06 & 71.89 \\
dog & 77.77 & 88.32 \\
frog & 71.41 & 82.16 \\
horse & 57.58 & 67.24 \\
ship & \textbf{50.38} & \textbf{59.83} \\
truck & 51.18 & 60.94 \\
\midrule
\textbf{Average} & \textbf{64.52} & \textbf{74.56} \\
\bottomrule
\end{tabular}
\end{table}

\begin{figure}[t]
\centering
\includegraphics[width=\columnwidth]{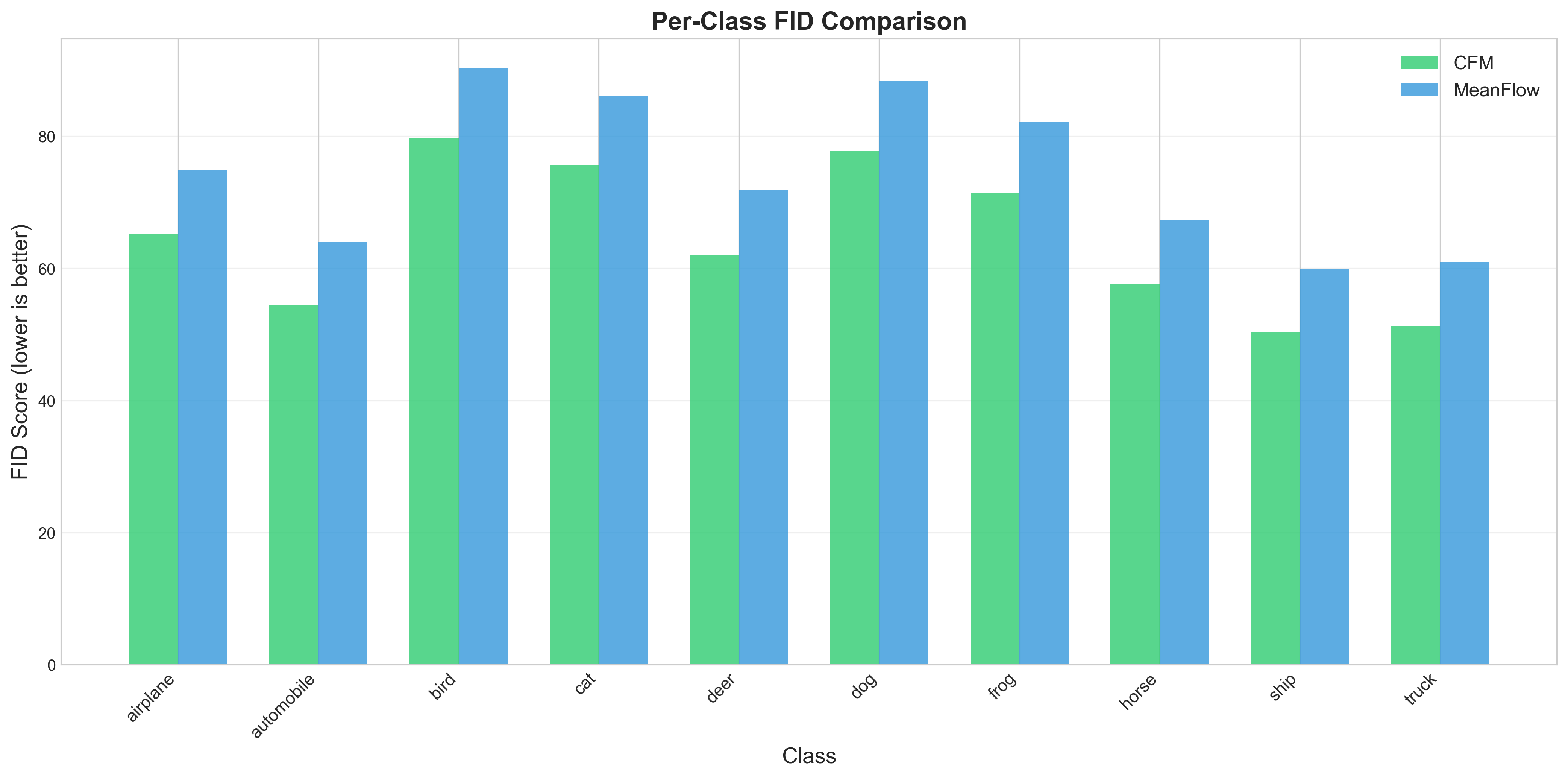}
\caption{Per-class FID comparison between CFM and MeanFlow. ``Ship'' achieves the best FID across both methods, while ``bird'' and ``dog'' are most challenging.}
\label{fig:per_class_fid}
\end{figure}

Fig.~\ref{fig:per_class_fid} and Table~\ref{tab:per_class_fid} show per-class performance:
\begin{itemize}
    \item \textbf{Best:} ``ship'' (CFM: 50.38, MeanFlow: 59.83) and ``truck'' (51.18, 60.94)
    \item \textbf{Worst:} ``bird'' (79.66, 90.23) and ``dog'' (77.77, 88.32)
    \item CFM consistently outperforms MeanFlow across all classes
\end{itemize}

\subsection{Inpainting Results}

\begin{figure*}[t]
\centering
\includegraphics[width=\textwidth]{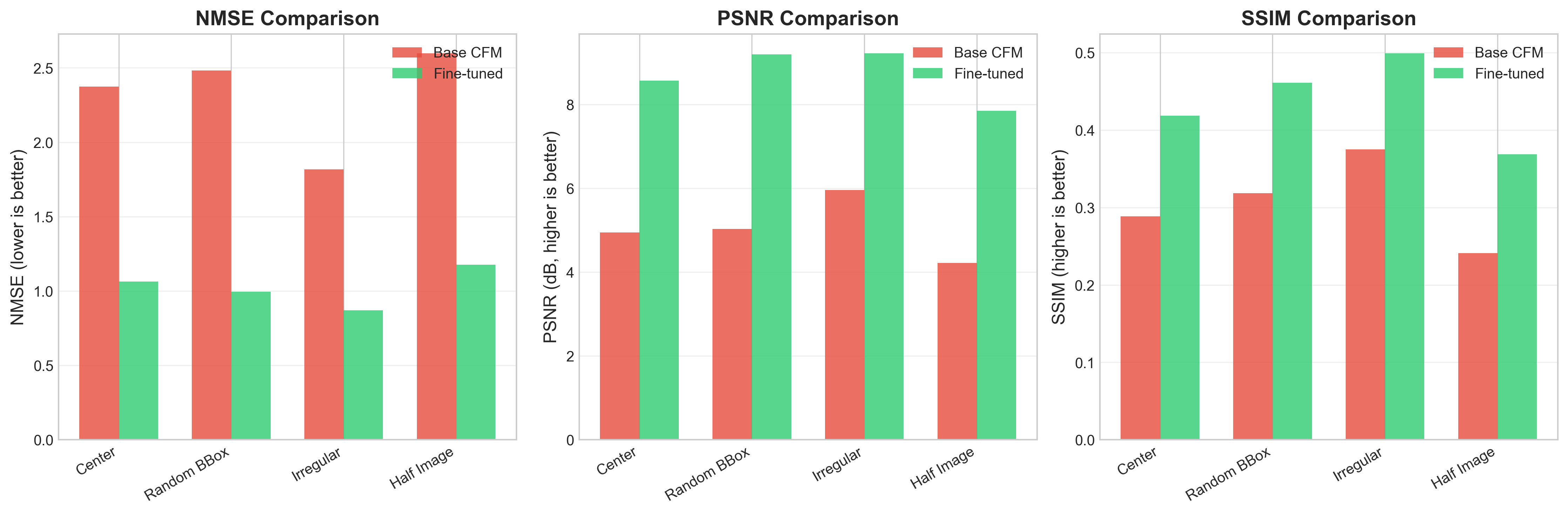}
\caption{Inpainting performance comparison: Base CFM (red) vs Fine-tuned (green) across all mask types. Fine-tuning dramatically improves all metrics---reducing NMSE by 52--60\% and improving PSNR by 55--86\%.}
\label{fig:inpaint_comparison}
\end{figure*}

\begin{table*}[t]
\centering
\caption{Comprehensive Inpainting Performance: Base vs Fine-tuned Model}
\label{tab:inpaint_comprehensive}
\begin{tabular}{l|ccc|ccc|ccc}
\toprule
& \multicolumn{3}{c|}{\textbf{NMSE}$\downarrow$} & \multicolumn{3}{c|}{\textbf{PSNR (dB)}$\uparrow$} & \multicolumn{3}{c}{\textbf{SSIM}$\uparrow$} \\
\textbf{Mask Type} & Base & Fine-tuned & $\Delta$\% & Base & Fine-tuned & $\Delta$\% & Base & Fine-tuned & $\Delta$\% \\
\midrule
Center & 2.37 & 1.06 & $-$55.2 & 4.9 & 8.6 & +73.2 & 0.289 & 0.418 & +44.9 \\
Random BBox & 2.48 & 1.00 & $-$59.9 & 5.0 & 9.2 & +82.8 & 0.318 & 0.461 & +44.9 \\
Irregular & 1.82 & 0.87 & $-$52.2 & 6.0 & 9.2 & +54.7 & 0.375 & 0.499 & +33.0 \\
Half Image & 2.60 & 1.18 & $-$54.7 & 4.2 & 7.8 & +86.0 & 0.241 & 0.369 & +53.0 \\
\midrule
\textbf{Average} & \textbf{2.32} & \textbf{1.03} & $\mathbf{-55.5}$ & \textbf{5.0} & \textbf{8.7} & $\mathbf{+74.2}$ & \textbf{0.306} & \textbf{0.437} & $\mathbf{+42.8}$ \\
\bottomrule
\end{tabular}
\end{table*}

Table~\ref{tab:inpaint_comprehensive} and Fig.~\ref{fig:inpaint_comparison} show the dramatic improvement from fine-tuning:
\begin{itemize}
    \item \textbf{NMSE:} Reduced by 52--60\% across all mask types
    \item \textbf{PSNR:} Improved by 55--86\%, with half-image showing the largest gain
    \item \textbf{SSIM:} Improved by 33--53\%, indicating better structural coherence
\end{itemize}

\begin{figure}[t]
\centering
\includegraphics[width=\columnwidth]{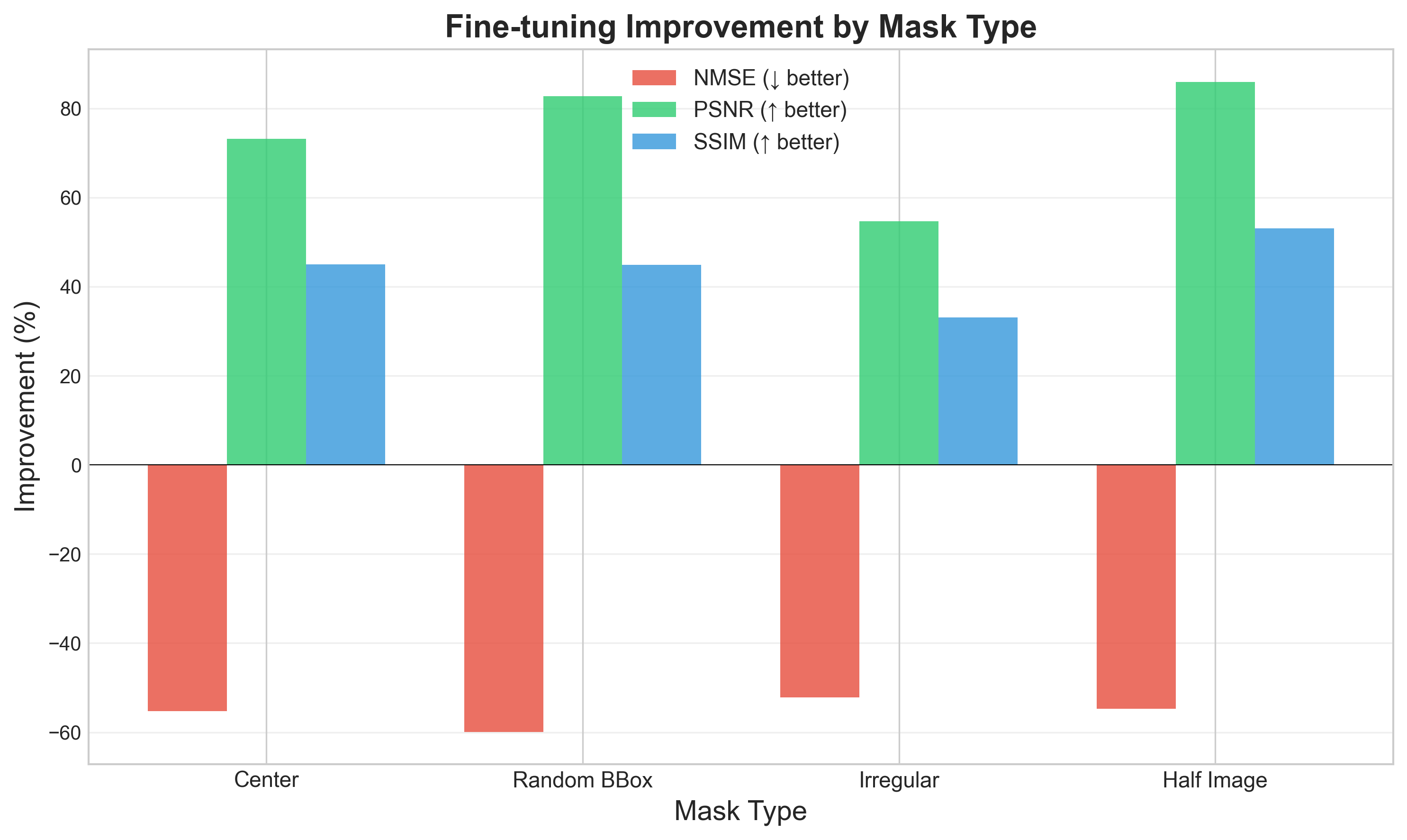}
\caption{Percentage improvement from fine-tuning by mask type. Half-image shows the largest PSNR improvement (+86\%), while random bbox shows the largest NMSE reduction ($-$60\%).}
\label{fig:inpaint_improvement}
\end{figure}

\subsection{Qualitative Inpainting Results}

\begin{figure*}[t]
\centering
\begin{subfigure}[b]{0.48\textwidth}
    \includegraphics[width=\textwidth]{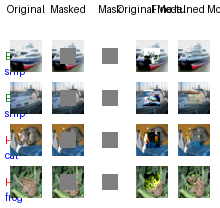}
    \caption{Center Mask}
\end{subfigure}
\hfill
\begin{subfigure}[b]{0.48\textwidth}
    \includegraphics[width=\textwidth]{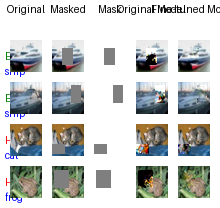}
    \caption{Random Bounding Box}
\end{subfigure}
\\[1em]
\begin{subfigure}[b]{0.48\textwidth}
    \includegraphics[width=\textwidth]{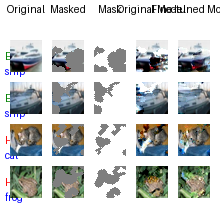}
    \caption{Irregular Brush Strokes}
\end{subfigure}
\hfill
\begin{subfigure}[b]{0.48\textwidth}
    \includegraphics[width=\textwidth]{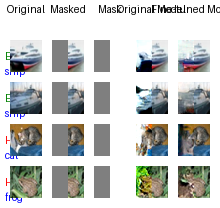}
    \caption{Half Image}
\end{subfigure}
\caption{Qualitative inpainting comparison across all four mask types. Each panel shows: Original $\rightarrow$ Masked $\rightarrow$ Base Model Result $\rightarrow$ Fine-tuned Model Result. The fine-tuned model produces more coherent completions with better color consistency and structural integrity. Note the improved ship reconstruction (row 1-2), better cat face completion (row 3), and more natural frog textures (row 4).}
\label{fig:inpaint_qualitative}
\end{figure*}

Fig.~\ref{fig:inpaint_qualitative} shows qualitative inpainting results across all mask types. Key observations:

\textbf{Center Mask:} The fine-tuned model reconstructs the central region with better color consistency. Ship images show cleaner hull reconstruction, and cat faces have more natural features.

\textbf{Random BBox:} Both models handle small random masks well, but the fine-tuned model shows better edge blending.

\textbf{Irregular Mask:} The scattered brush stroke pattern is challenging. The fine-tuned model produces smoother completions with fewer artifacts.

\textbf{Half Image:} The most challenging case. While neither model perfectly reconstructs missing halves, the fine-tuned model generates more plausible content with better texture matching.

\subsection{Per-Class Inpainting Analysis}

\begin{figure}[t]
\centering
\includegraphics[width=\columnwidth]{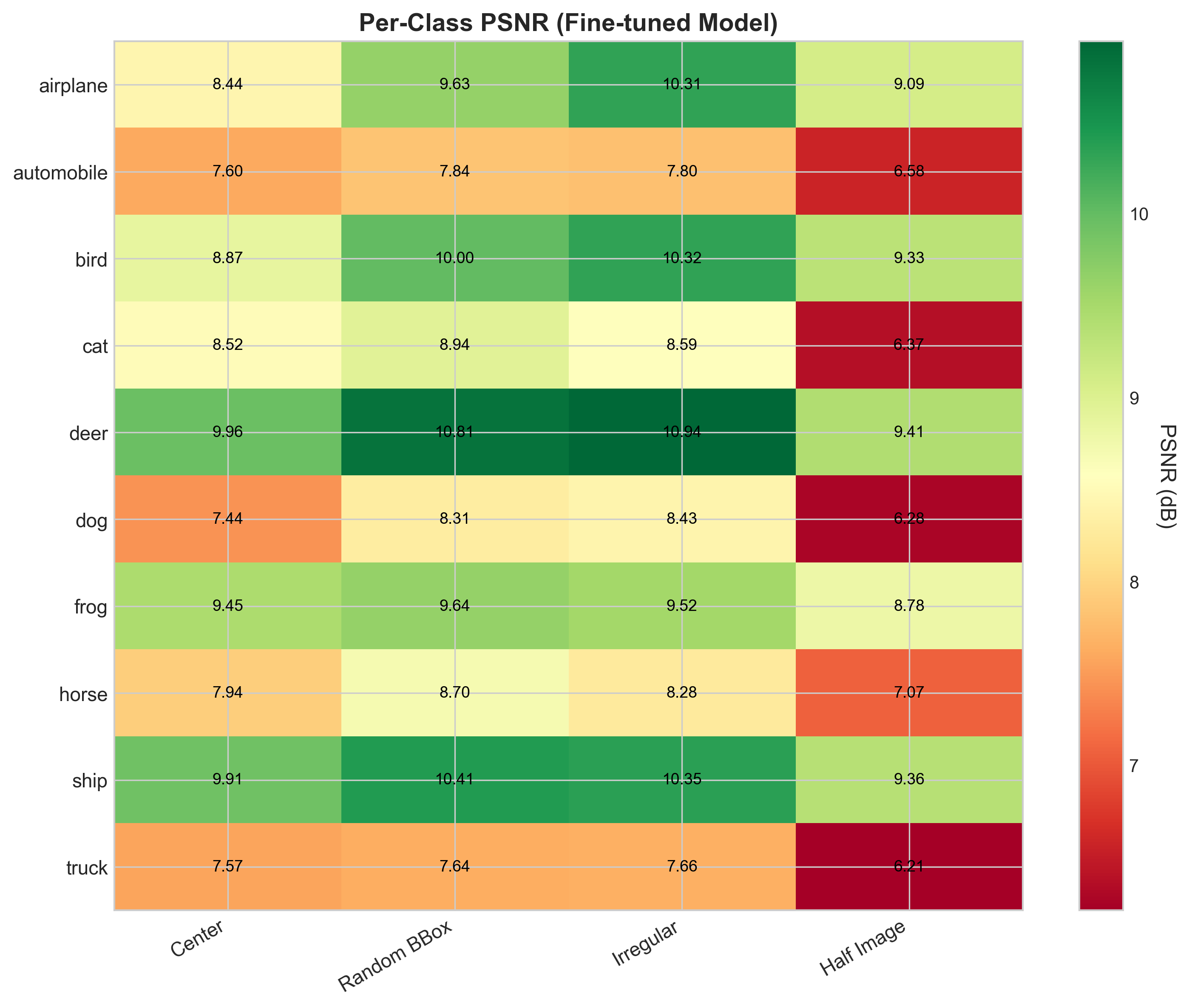}
\caption{Per-class PSNR heatmap for the fine-tuned model. ``Deer'' consistently achieves the highest PSNR across all mask types (9.41--10.94 dB), while ``truck'' and ``dog'' are most challenging.}
\label{fig:psnr_heatmap}
\end{figure}

\begin{table}[t]
\centering
\caption{Fine-tuned Model: Center Mask Per-Class Performance}
\label{tab:ft_center}
\begin{tabular}{lccc}
\toprule
\textbf{Class} & \textbf{NMSE}$\downarrow$ & \textbf{PSNR}$\uparrow$ & \textbf{SSIM}$\uparrow$ \\
\midrule
airplane & 1.006 & 8.44 & 0.391 \\
automobile & 1.050 & 7.60 & 0.371 \\
bird & 1.149 & 8.87 & 0.405 \\
cat & 1.158 & 8.52 & 0.398 \\
deer & 1.003 & 9.96 & 0.504 \\
dog & 1.226 & 7.44 & 0.388 \\
frog & 1.032 & 9.45 & 0.436 \\
horse & 1.075 & 7.94 & 0.408 \\
ship & \textbf{0.814} & \textbf{9.91} & 0.462 \\
truck & 1.120 & 7.57 & 0.420 \\
\midrule
\textbf{Overall} & \textbf{1.063} & \textbf{8.57} & \textbf{0.418} \\
\bottomrule
\end{tabular}
\end{table}

\begin{table}[t]
\centering
\caption{Fine-tuned Model: Half Image Per-Class Performance}
\label{tab:ft_half}
\begin{tabular}{lccc}
\toprule
\textbf{Class} & \textbf{NMSE}$\downarrow$ & \textbf{PSNR}$\uparrow$ & \textbf{SSIM}$\uparrow$ \\
\midrule
airplane & \textbf{0.876} & 9.09 & 0.472 \\
automobile & 1.176 & 6.58 & 0.382 \\
bird & 1.053 & 9.33 & 0.429 \\
cat & 1.567 & 6.37 & 0.335 \\
deer & 1.112 & \textbf{9.41} & 0.420 \\
dog & 1.582 & 6.28 & 0.256 \\
frog & 1.044 & 8.78 & \textbf{0.512} \\
horse & 1.232 & 7.07 & 0.257 \\
ship & 0.849 & 9.36 & 0.372 \\
truck & 1.265 & 6.21 & 0.252 \\
\midrule
\textbf{Overall} & \textbf{1.176} & \textbf{7.85} & \textbf{0.369} \\
\bottomrule
\end{tabular}
\end{table}

\begin{figure}[t]
\centering
\includegraphics[width=\columnwidth]{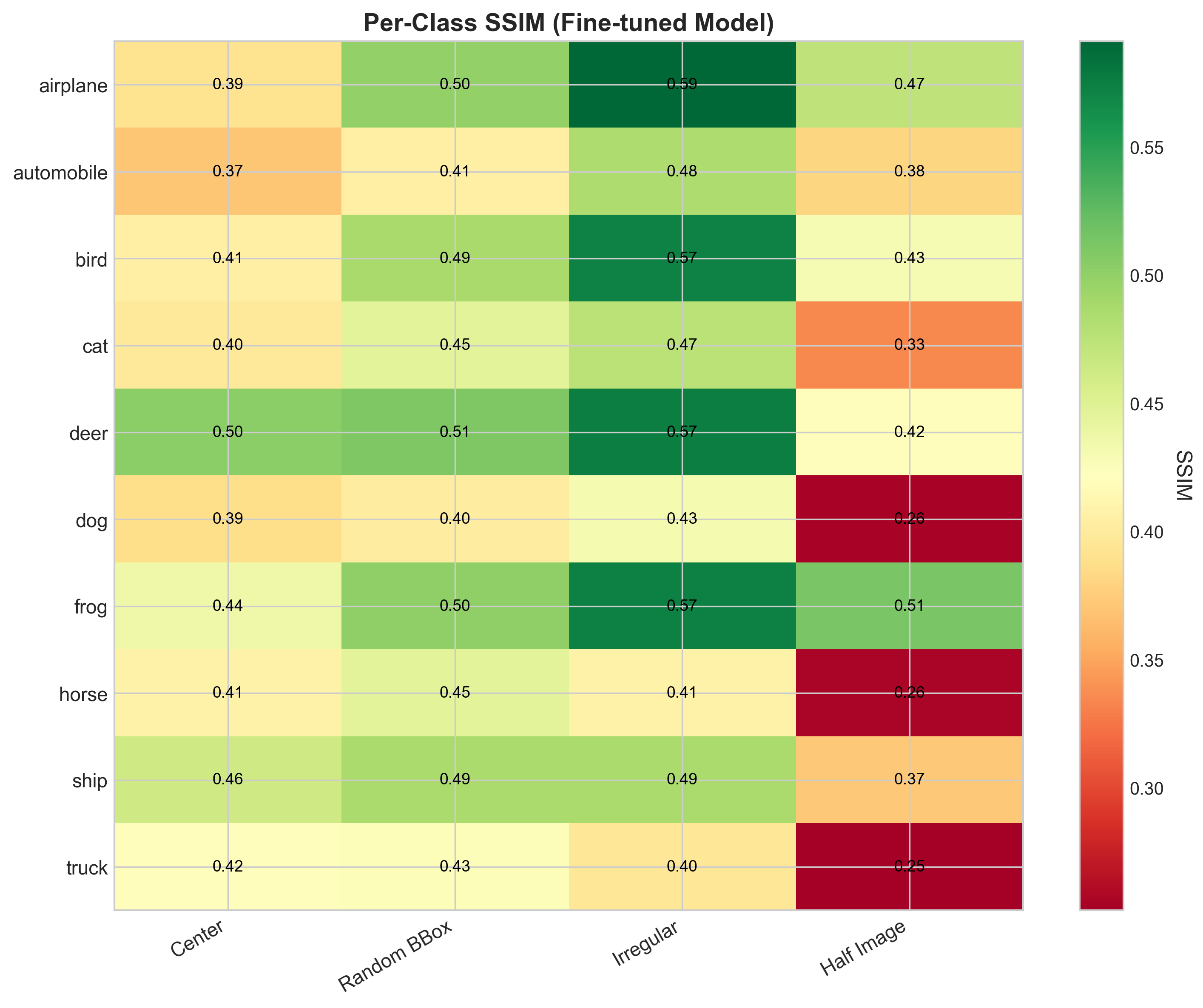}
\caption{Per-class SSIM heatmap for the fine-tuned model. Higher values (green) indicate better structural similarity. ``Airplane'' and ``frog'' achieve high SSIM on irregular masks.}
\label{fig:ssim_heatmap}
\end{figure}

\subsection{Mask Difficulty Analysis}

\begin{figure}[t]
\centering
\includegraphics[width=\columnwidth]{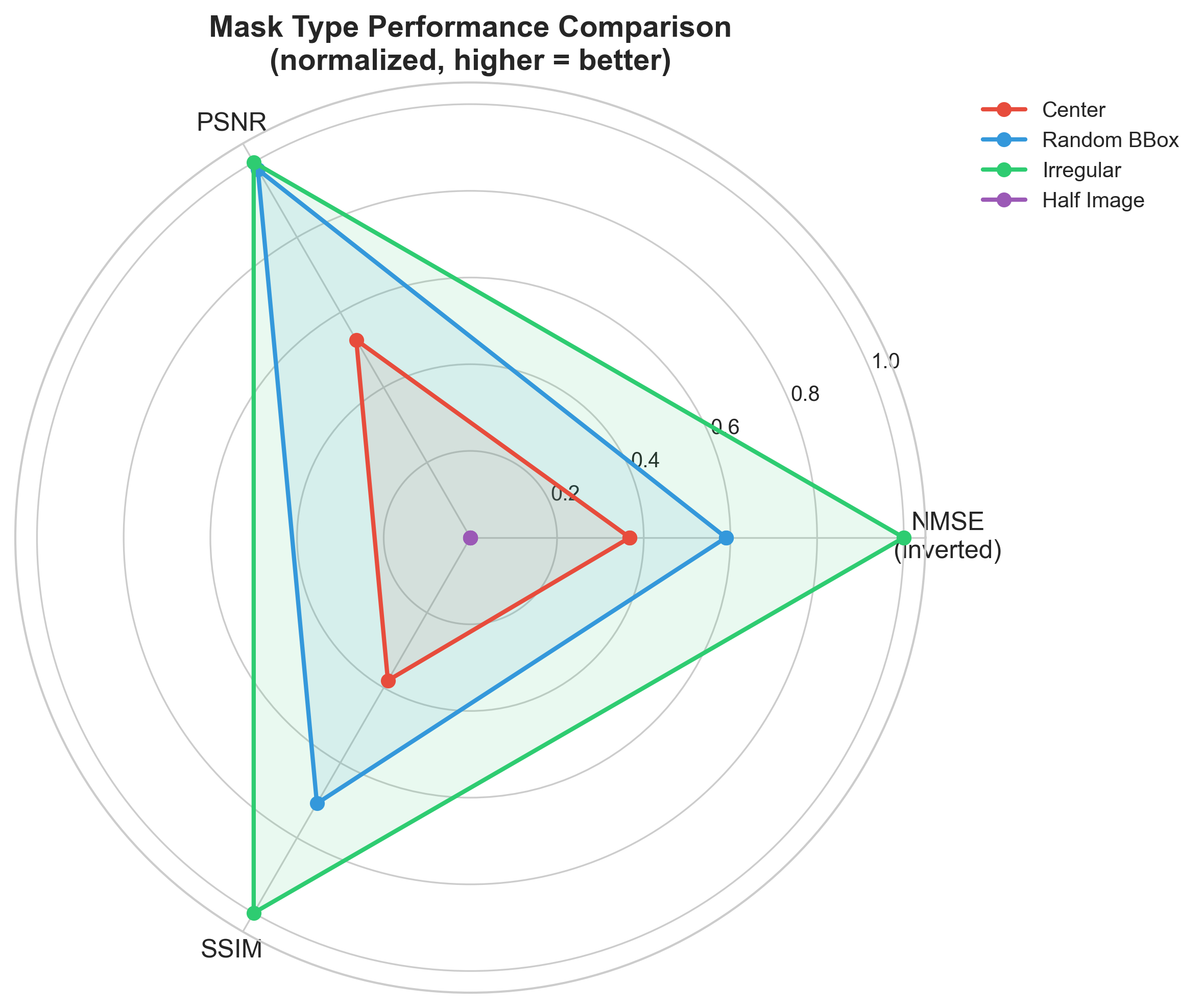}
\caption{Normalized performance comparison across mask types. Irregular masks (green) achieve the best overall performance, while half-image (purple) is most challenging. Metrics are normalized so higher values indicate better performance.}
\label{fig:mask_radar}
\end{figure}

Fig.~\ref{fig:mask_radar} shows the relative difficulty of each mask type:
\begin{itemize}
    \item \textbf{Easiest:} Irregular masks achieve best scores across all metrics
    \item \textbf{Moderate:} Random BBox and Center masks show similar difficulty
    \item \textbf{Hardest:} Half-image completion is most challenging due to missing 50\% of context
\end{itemize}

\subsection{Per-Class Patterns}
From Figs.~\ref{fig:psnr_heatmap} and \ref{fig:ssim_heatmap}:
\begin{itemize}
    \item \textbf{Best classes:} ``deer'' (PSNR: 9.41--10.94), ``ship'' (9.36--10.41), ``frog'' (8.78--9.64)
    \item \textbf{Challenging classes:} ``truck'' (6.21--7.66), ``dog'' (6.28--8.43), ``automobile'' (6.58--7.84)
    \item Pattern mirrors generation quality: simpler geometry $\Rightarrow$ better inpainting
\end{itemize}

\section{Discussion}
\label{sec:discussion}

\textbf{Why DDPM Underperforms.} As shown in Fig.~\ref{fig:ddpm_failure}, DDPM fails to generate coherent images even after 400 epochs. This is due to: (1) insufficient timesteps ($T=200$ vs typical 1000), (2) the TinyUNet architecture being too small for effective noise prediction, and (3) the inherent difficulty of $\epsilon$-prediction compared to velocity matching.

\textbf{CFM vs MeanFlow Trade-offs.} CFM achieves better FID (24.15 vs 29.15) but requires 50 ODE steps. MeanFlow sacrifices 5 FID points for 50$\times$ faster inference, making it suitable for real-time applications.

\textbf{Fine-tuning Effectiveness.} The dramatic improvement (74\% average PSNR gain) demonstrates that:
\begin{enumerate}
    \item Base generation models are not optimal for inpainting
    \item Training with masked inputs helps boundary harmonization
    \item The combined loss balances global coherence with local reconstruction
\end{enumerate}

\textbf{Class-Dependent Performance.} Classes with simpler, more uniform textures (deer, ship, frog) achieve better inpainting, while complex textures with fine details (dog, truck) remain challenging. This pattern is consistent between generation and inpainting tasks.

\section{Related Work}
\label{sec:related}

\textbf{Diffusion Models.} DDPM \cite{ho2020denoising} and score-based models \cite{song2020score} established diffusion as a leading paradigm. DDIM \cite{song2020denoising} enabled faster sampling.

\textbf{Flow Matching.} Lipman et al. \cite{lipman2023flow} introduced flow matching. Rectified flows \cite{liu2023flow} improved efficiency through straighter paths.

\textbf{One-Step Methods.} Consistency models \cite{song2023consistency} and progressive distillation \cite{salimans2022progressive} reduce steps but require pre-trained teachers. MeanFlow \cite{meanflow2025} achieves one-step generation without distillation.

\textbf{Image Inpainting.} Traditional approaches use GANs \cite{yu2018generative} or partial convolutions \cite{liu2018image}. Recent work applies diffusion with RePaint \cite{lugmayr2022repaint} strategies.

\section{Conclusion}
\label{sec:conclusion}

We presented a comprehensive comparison of DDPM, CFM, and MeanFlow on CIFAR-10, with an extension to image inpainting. Key findings:

\begin{enumerate}
    \item \textbf{CFM significantly outperforms DDPM} (FID 24.15 vs 402.98) with the same architecture
    \item \textbf{MeanFlow enables one-step generation} (FID 29.15) with 50$\times$ speedup
    \item \textbf{Fine-tuning dramatically improves inpainting}: +74\% PSNR, +43\% SSIM
    \item \textbf{Irregular masks} are easiest, \textbf{half-image} is hardest
    \item \textbf{``Deer'' and ``ship''} achieve best results; ``dog'' and ``truck'' are most challenging
\end{enumerate}

Future work includes scaling to higher resolutions, combining MeanFlow with latent diffusion, and exploring progressive inpainting strategies.


\bibliographystyle{IEEEtran}
\bibliography{name}

\end{document}